\documentclass[twoside]{article} \usepackage{aistats2017}

\usepackage[pdftex]{graphicx} 
\usepackage{amsfonts,amsmath,amssymb,amsthm,mathtools} 
\usepackage{wrapfig}  
\usepackage{mdwlist}  
\usepackage{sidecap}  
\usepackage{bbm}
\usepackage{bm,amsbsy} 
\usepackage{algorithm,algorithmic}
\usepackage{mathtools}
\usepackage{multirow}
\usepackage{graphicx}

\usepackage[font=small,labelfont=bf]{caption}

\newcommand\myeq{\stackrel{\mathclap{\normalfont\mbox{def}}}{=}}

\usepackage{url}            

\newcommand{\figref}[1]{Fig.~\ref{fig:#1}}  
\newcommand{\secref}[1]{Sec.~\ref{#1}}  

\newcommand{\vx}{\mathbf{x}}

\newcommand{\Dat}{\mathcal{D}}

\newcommand{\valpha}{\mathbf{\ensuremath{\bm{\alpha}}}}

\newcommand{\vz}{\mathbf{z}}

\newcommand{\vpi}{\mathbf{\ensuremath{\bm{\pi}}}}
\newcommand{\vmu}{\mathbf{\ensuremath{\bm{\mu}}}}

\newcommand{\vtheta}{\mathbf{\ensuremath{\bm{\theta}}}}

\newcommand{\Nrm}{\mathcal{N}}
\newcommand{\trp}{{^\top}} 

\renewcommand{\eqref}[1]{Eq~(\ref{eq:#1})}

\usepackage{color}  
\definecolor{gray}{rgb}{0.5, .5, .5}

\begin{document}

%

%

\twocolumn[

\aistatstitle{DP-EM: Differentially Private Expectation Maximization}

\aistatsauthor{Mijung Park
\And Jimmy Foulds
\And Kamalika Chaudhuri
\And Max Welling}
%
\aistatsaddress{University of Amsterdam
\And 
UC, San Diego
\And
UC, San Diego
\And
University of Amsterdam} 

]

\begin{abstract}

The iterative nature of the expectation maximization (EM) algorithm presents a challenge for privacy-preserving estimation, as each iteration increases the amount of noise needed.
We propose a \emph{practical} private EM algorithm that overcomes this challenge using two innovations: (1) a novel moment perturbation formulation for differentially private EM (DP-EM), and (2) the use of two recently developed composition methods to bound the privacy ``cost'' of multiple EM iterations: the \emph{moments accountant (MA)} and \emph{zero-mean concentrated differential privacy (zCDP)}.  Both MA and zCDP bound the moment generating function of the privacy loss random variable and achieve a refined tail bound, which effectively decrease the amount of additive noise. 
We present empirical results showing the benefits of our approach, as well as similar performance between these two composition methods in the DP-EM setting for Gaussian mixture models.
Our approach can be readily extended to many iterative learning algorithms,  opening up various exciting future directions. 
\end{abstract}


\section{Introduction}

Data on all aspects of our daily lives, such as behavioural, health and financial data, are increasingly collected, stored and analyzed by corporations and government agencies, and there is a dire need for developing machine learning tools that can analyze these data while still guaranteeing the privacy of individuals.  Much progress has been made recently in developing privacy-preserving methods~\cite{Dwork14, SC13} and \emph{differential privacy}, in particular, is emerging as the dominant notion of algorithmic privacy \cite{Dwork14}.

In this paper, we derive differentially private variants of the expectation maximization (EM) algorithm which has been widely used to solve statistical problems in many areas of science including bioinformatics \cite{bailey1994fitting}, neuroscience \cite{zhang2001segmentation}, and computer vision \cite{carson2002blobworld}. Expectation maximization iteratively estimates the parameters of models with unobserved variables. We present a very general privacy-preserving EM algorithm which can be used for any model with a complete-data likelihood in the exponential family. We then apply our algorithm to the mixture of Gaussians (MoG) density estimation model and the factor analysis (FA) model. Having access to a private density estimator is particularly valuable because it provides a means to anonymize the data in a principled way, by simply sampling a dataset from the model and replacing the original data with this sampled data.  

Since differentially private machine learning algorithms usually achieve privacy by adding noise to perturb the output of the algorithm or its intermediate stages, the main challenge in developing privacy-preserving algorithms is in controlling the associated loss in {\em{statistical efficiency}} or utility per sample. This problem is particularly exacerbated for iterative algorithms such as EM. For example, recent work on the $k$-means algorithm, a variant of EM for mixture of Gaussians, requires adding noise to the parameters where the noise standard deviation is on the order of the input dimension \emph{times the number of iterations} \cite{Blum:2005:PPS:1065167.1065184}, which may necessitate early termination. To avoid this, more recent work proposes to apply a standard $k$-means clustering algorithm to a privatized synopsis of the data \cite{Su2016}. Their synopsis generation method consists of putting rectangular bounding boxes in the data space and counting how many data points are in each box. However, this method applies mainly to the clustering task and for low-dimensional data. 

Instead, we propose to resolve the privacy-utility dilemma using two key innovations: a private EM formulation based on moment perturbation for sensible use of the privacy budget \emph{per iteration}, and recently proposed composition methods to improve the privacy cost \emph{across many iterations}. Our moment perturbation approach is applicable for any model in which the complete-data likelihood is in the exponential family. In such cases, the EM parameters are functions of moments of latent and observed variables, which we perturb for privacy. Moment perturbation for differentially private estimators is not a new concept (see \cite{foulds:bayesianPrivacy16,Smith08}). However, unlike \cite{Smith08}, we do not require subsampling of the data.

Furthermore, our algorithm calculates the cumulative privacy cost using two refined composition methods, the \emph{moments accountant} and \emph{zCDP}.  
 The moments accountant \cite{2016arXiv160700133A} bounds the moments of the privacy loss random variable. Inspired by CDP \cite{2016arXiv160301887D}, zCDP \cite{BunS16}  formulates the moments of the privacy loss random variable in terms of the R{\'e}nyi  divergence between the output distributions obtained by running an algorithm on two datasets that differ in the value of a single individual. In both cases, the moments bound yields a tighter tail bound, and consequently, for a given total privacy budget, allows for a higher per-iteration budget than standard methods.
%
%
Our experimental results show that by combining our moment perturbation formulation of privacy-preserving EM with refined composition methods, we obtain a practical and effective algorithm for privately estimating the parameters of latent variable models.

We start by reviewing differential privacy, the moments accountant, and EM in \secref{EMsummary}. In \secref{DPEMgeneral}, we introduce our general DP-EM framework.
We then derive the DP-EM algorithm  for mixture of Gaussians in \secref{DPEMMoG}.
In \secref{CDPEM}, we construct the MA and zCDP formulation for EM under MoGs.  In \secref{FA}, we provide the DP-EM algorithm for factor analysis, 
and we illustrate the effectiveness of our algorithms in \secref{Illustration}.


\section{Background}\label{EMsummary}
\vspace{-0.2cm}
In this section, we provide background information on the definitions of algorithmic privacy that we use, the MA and zCDP formulations which provide a refined privacy analysis, as well as the general EM algorithm. 
\vspace{-0.2cm}
\paragraph{Differential privacy.} Differential privacy (DP) is a formal definition of the privacy properties of data analysis algorithms \cite{Dwork14}.  
Given an algorithm $\mathcal{M}$ and datasets $\mathbf{X}$, $\mathbf{X}'$ differing by a single entry,
the \emph{privacy loss} random variable of an outcome $o$ is
\vspace{-0.1cm}
\begin{equation}
L^{(o)} = \log \frac{Pr(\mathcal{M}_{(\mathbf{X})} = o)}{Pr(\mathcal{M}_{(\mathbf{X}')} = o)} \mbox{ .}
\vspace{-0.1cm}
\end{equation} 
$\mathcal{M}$ is $\epsilon$-DP if and only if
$|L^{(o)}| \leq \epsilon, \forall o$.
Intuitively, the definition states that the output probabilities must not change very much when a single individual's data is modified, thereby limiting the amount of information that the algorithm reveals about any one individual.
An approximate version is ($\epsilon, \delta$)-DP, defined to hold if and only if
$|L^{(o)}| \leq \epsilon$, with probability at least $1-\delta$.

\paragraph{Concentrated differential privacy.} 
Concentrated differential privacy (CDP) is a recently proposed relaxation of differential privacy which aims to make privacy-preserving iterative algorithms more practical than for DP while still providing strong privacy guarantees. 
There are two variants of CDP. 
First, in {\it{$(\mu, \tau)$-mCDP}} \cite{2016arXiv160301887D}, $L^{(o)}$ subtracted by its mean $\mu$ is subgaussian with standard deviation $\tau$:  $E[e^{\lambda (L^{(o)}-\mu) }] \leq e^{{\lambda^2 \tau^2}/{2}}, \forall \lambda \in \mathbb{R}$. 
Second, in {\it{$\tau$-zCDP}} \cite{BunS16}, that arises from a connection between the moment generating function of $L^{(o)}$ and the R{\'e}nyi divergence between the distributions of $\mathcal{M}_{(\mathbf{X})}$ and that of $\mathcal{M}_{(\mathbf{X}')}$, we require: $ e^{(\alpha-1)\mbox{D}_\alpha} = E[e^{(\alpha-1) L^{(o)} }] \leq e^{(\alpha-1)\alpha \tau}, \forall \alpha \in (1, \infty)$,
where
the $\alpha$-R{\'e}nyi divergence is denoted by $\mbox{D}_\alpha = \mbox{D}_\alpha(Pr(\mathcal{M}_{(\mathbf{X})})|| Pr(\mathcal{M}_{(\mathbf{X'})}))$. Observe that in this case $L^{(o)}$ is also subgaussian but zero-mean.
In zCDP, composition is straightfoward since the R{\'e}nyi divergence between two product distributions is simply the sum of the R{\'e}nyi divergences of the marginals.  

%
%

We will use zCDP rather than mCDP, since many DP and approximate DP mechanisms can be characterised in terms of zCDP, but not in terms of mCDP without a large loss in privacy parameters.  This correspondence will allow us to use zCDP as a tool for analyzing \emph{composition under the $(\epsilon,\delta)$-DP privacy definition}, for a fair comparison between CDP and DP analyses.\footnote{See Sec. 4 in \cite{BunS16} for a detailed explanation.}

\vspace{-0.2cm}

\paragraph{Moments accountant.} 
The moments accountant calculates a privacy budget by bounding the moments of $L^{(o)}$, where 
the $\lambda$-th moment is defined as the log of the moment generating function evaluated at $\lambda$ \cite{2016arXiv160700133A}:
\vspace{-0.1cm}
\begin{equation}
\label{eq:Moment}
\alpha_{\mathcal{M}}(\lambda; \Dat, \Dat') = \log \mathbb{E}_{o \sim \mathcal{M}(\Dat)} \left[ e^{\lambda  L^{(o)}}\right].
\vspace{-0.1cm}
\end{equation} 
The worst case over all the neighbouring databases  $\alpha_{\mathcal{M}}(\lambda)$ is defined as 
$ \alpha_{\mathcal{M}}(\lambda) = \max_{\Dat, \Dat'}\alpha_{\mathcal{M}}(\lambda; \Dat, \Dat').$\footnote{The form of $\alpha_{\mathcal{M}}(\lambda)$ is determined by the mechanism.}

Using Markov's inequality, for any $\epsilon>0$, the $\lambda$-th moment is converted to the ($\epsilon,\delta$)-DP guarantee by\footnote{See Appendix A in \cite{2016arXiv160700133A} for the proof.}
\vspace{-0.1cm}
\begin{equation}
\label{eq:tail_bound}
\delta = \min_{\lambda} \exp \left[ \alpha_{\mathcal{M}}(\lambda) - \lambda \epsilon \right].
\vspace{-0.1cm}
\end{equation} 
The $\lambda$-th moment in \eqref{Moment} composes linearly, which yields the composability theorem (Theorem 2.1 in \cite{2016arXiv160700133A}). An immediate result from the composibility theorem is that the sum of each upper bound on $\alpha_{\mathcal{M}_j}$ is an upper bound on the total $\lambda$th moment after $J$ compositions, 
\vspace{-0.2cm}
\begin{equation}
\label{eq:Composibility}
\alpha_{\mathcal{M}}(\lambda) \leq  \sum_{j=1}^J \alpha_{\mathcal{M}_j}(\lambda). 
\vspace{-0.2cm}
\end{equation} 
%


\paragraph{The general EM algorithm.}
Given $N$ \emph{i.i.d.} observations $X := \{\vx_i\}_{i=1}^N$, with each observation $\vx_i \in \mathbb{R}^d$, and hidden variables $Z := \{\vz_i\}_{i=1}^N$, computing the maximum likelihood estimator of a vector of model parameters $\vtheta = [\theta_1, \cdots, \theta_L]$ is analytically intractable, due to the integral or summation inside the logarithm,
\begin{equation}
\mathcal{L}(\vtheta)=\log p(X | \vtheta) =  \log \int dZ \; p(X, Z | \vtheta).
\end{equation} 
%
Instead, one can lower-bound $\mathcal{L}(\vtheta)$ using the posterior distribution over latent variables $q(Z)$ \cite{neal1998view},
\vspace{-0.3cm}
\begin{equation}
\mathcal{L}(\vtheta) \geq 
\int dZ \; q(Z)  \log  \tfrac{ p(X, Z| \vtheta)}{q(Z)} \quad \myeq \quad \mathcal{F}(q, \vtheta),
\vspace{-0.3cm}
\end{equation} where the lower bound is often called {\it{free energy}} \cite{Feynman}, 
%
 $\mathcal{F}(q, \vtheta) = \langle \log p(X, Z|\vtheta) \rangle_{q(Z)} + H(q)$,
where $H(q)$ is the entropy of $q(Z)$.
%
EM alternates between: 
(1) the E-step: optimizing $ \mathcal{F}$ wrt distribution over unobserved variables holding parameters fixed
 \begin{equation}
 q^{(j)}(Z) = \arg\max_{q(Z)} \mathcal{F}(q(Z), \vtheta^{(j-1)})
\end{equation} and 
(2) the M-step: maximizing $ \mathcal{F}$ wrt parameters holding  the latent distribution fixed
 \begin{equation}\label{eq:MLE}
\vtheta^{(j)} 
= \arg \max_{\vtheta} \mathcal{F}(q^{(j)}(Z), \vtheta)
\end{equation} where $\mathcal{F}(q^{(j)}(Z), \vtheta) = \langle \log p(X, Z|\vtheta) \rangle_{q^{(j)}(Z)} + \mbox{const}$ since $H(q)$ does not directly depend on $\vtheta$. 

To understand what EM does, one can rewrite the free energy in terms of the log-likelihood and the KL divergence terms, 
$\mathcal{F}(q, \vtheta) = \mathcal{L}(\vtheta)- D_{KL}\left[ q(Z) || p(Z|X, \vtheta) \right].$
During the E-step, we set $ q^{(j)}(Z) =  p(Z|X, \vtheta^{j-1})$, which makes the second term zero and the free energy equals the likelihood. Then, in the M-step, we get the maximum likelihood estimate (MLE). For the maximum a posteriori (MAP) estimate, we add the log prior for the parameters $\log p(\vtheta)$ to the right hand side of \eqref{MLE}. 


\section{The general DP-EM algorithm}\label{DPEMgeneral}
The EM algorithm is frequently used for models whose joint distribution over observed and unobserved variables remains in the exponential family: $p(X, Z) = h(X, Z) \exp(\vtheta\trp T(X,Z))/A(\vtheta)$,  while the marginal $p(X)$ does not. In this case, the free energy can be rewritten as 
\begin{equation}
 \mathcal{F}(q, \vtheta) = \vtheta \trp \langle T(X, Z) \rangle_{q(Z)} - N \log A(\vtheta) + c,
\end{equation} where $c$ is some constant wrt $\vtheta$, and $\vtheta \trp \langle T(X, Z) \rangle_{q(Z)} = \sum_{i=1}^N \mathbb{E}_{q(\vz_i)}  \sum_{l=1}^L \theta_l T_l(\vx_i, \vz_i)$. In the E-step, we compute the expected sufficient statistics under $q$, i.e., $\langle T(X, Z) \rangle_{q(Z)}$. Then, in the M-step, we compute partial derivatives wrt each parameter, 
\vspace{-0.3cm}
 \begin{equation}\label{eq:paramupdate}
 \tfrac{\partial}{\partial \theta_l}  \mathcal{F} (q, \vtheta)  = 
\tfrac{1}{N}\sum_{i=1}^N \mathbb{E}_{q(\vz_i)}  T_l(\vx_i, \vz_i) - \tfrac{\partial}{\partial \theta_l} \log A(\vtheta) = 0.\nonumber
\vspace{-0.3cm}
 \end{equation}
Although it is not straightforward to derive a closed-form expression for each parameter update due to the dependence on other parameters in $A(\vtheta)$, it is easy to see that each parameter update depends on each expected sufficient statistics, i.e., moments, denoted by $M_l = \tfrac{1}{N}\sum_{i=1}^N \mathbb{E}_{q(\vz_i)}  T_l(\vx_i, \vz_i) $. So, 
to output privatized parameters, all we need is to perturb the moments to compensate any single data point's change. The sensitivity of the expected sufficient statistics is given by
\vspace{-0.2cm}
\begin{align}\label{eq:moment_sen}
& \Delta M_l  \\
 &= \max_{|\Dat-\tilde{\Dat}|_1 = 1} | M_l(\Dat) - \tilde{M}_l(\tilde{\Dat})|, \nonumber \\
  &= \max_{\vx_j, {\vx}'_j}  \tfrac{1}{N} |\mathbb{E}_{q(\vz_j)}  T_l(\vx_j, \vz_j)-   \mathbb{E}_{q({\vz}'_j)}  T_l({\vx}'_j, {\vz}'_j)| , \nonumber \\
& \leq \max_{\vx_j, {\vx}'_j} \tfrac{1}{N}  |\langle T_l(\vx_j, \vz_j) \rangle_{q(\vz_j)} | + \tfrac{1}{N}  |\langle T_l({\vx}'_j, {\vz}'_j) \rangle_{q({\vz}'_j)} |, \nonumber
\end{align} 
where the last line is due to the triangle inequality. 
The expectation over $\vz$ can be rewritten as an inner product, and using H\"{o}lder's inequality: 
$|\langle T_l(\vx_j, \vz_j) \rangle_{{q(\vz_j)}} |  = |\langle q(\vz_j), T_l(\vx_j, \vz_j) \rangle| 
 \leq |q(\vz_j) |_1 | T_l(\vx_j, \vz_j)|_{\infty}, $
where $|q(\vz_j) |_1 = 1$ and $|T_l(\vx_j, \vz_j)|_{\infty}$ is maximum over all $(\vx_j, \vz_j)$. 
As in many existing works (e.g., \cite{ERM, Kifer12privateconvex} among many others), we also assume that datasets are pre-processed such that the $L_2$ norm of any $\vx_i$ is less than $1$, meaning that any $\vx_i$ stays within a unit ball. 
Furthermore, we assume that $q(Z)$ has a bounded support of $Z$ denoted by $\mathcal{Z}$. Under these assumptions,  
the sensitivity is given by  
$\Delta M_l =  \max_{(\vx_j, \vz_j) \in (B_1(\mathcal{X}), \; \mathcal{Z})} \tfrac{2}{N} |  T_l(\vx_j, \vz_j) |. $
Using this sensitivity, we add noise to each moment and the perturbed moments are mapped by a model-specific deterministic function $g$ to the vector of privatized parameters, given as
$\tilde{\vtheta}^* = g(\{ \tilde{M}_l\}_{l=1, \cdots, L} ), $
 where $\tilde{M}_{l=1, \cdots, L}$ are perturbed moments.
%
Using this general framework, we derive the differentially private EM algorithm for mixture of Gaussians and factor analysis in the following.

\begin{figure*}[t]
\centering
\centerline{\includegraphics[width=\textwidth]{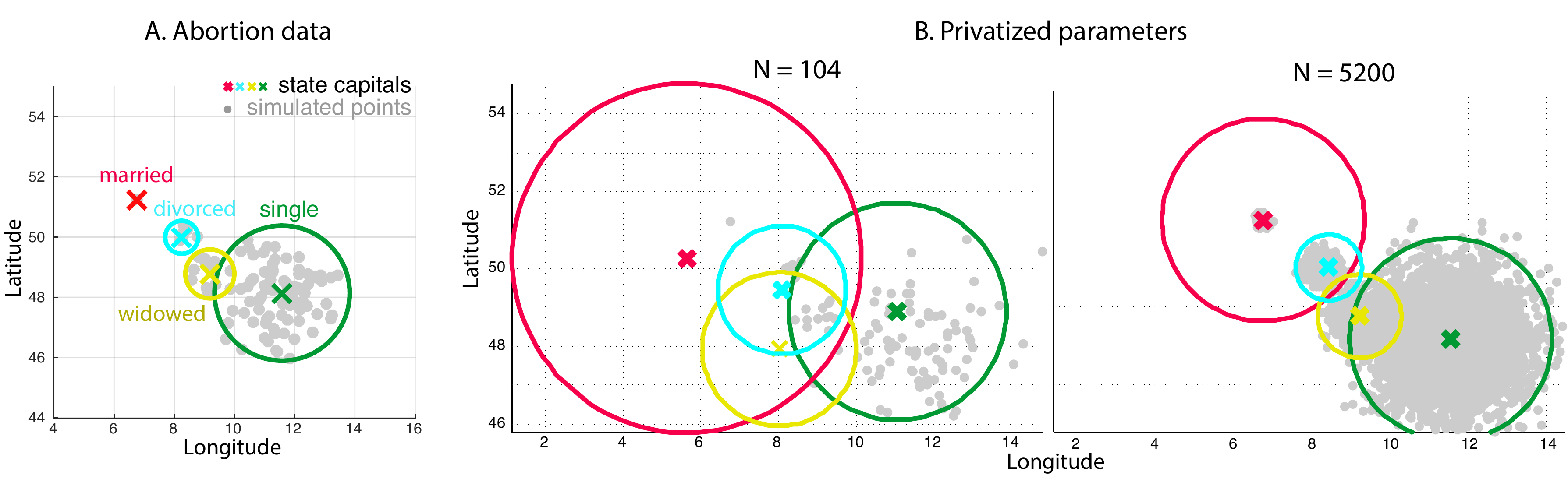}}
  \caption{ \textbf{A.} The abortion dataset (from $destatis.de$) provides 
  per-marital-status abortion rates 
  occurred in the state of Baden-W\"{u}rttemberg in 2015, as well as from which state each individual came from. Due to the lack of exact location, we simulated $104$ data points based on the abortion rate in each state (in grey). Notice that there is only one person who is originally from the state of North Rhine-Westphalia (top left, in red) and falls into the `married' category. Hence, the person's information is completely revealed in the mean parameter if one runs the conventional EM algorithm.     
  \textbf{B (Left).} Given the $104$ data points, by privatizing the mean and variance parameters as illustrated in \secref{Illustration}, the married person's information (top left, in red) is now not easily inferrable.    \textbf{B (Right).} When we have 50 times more datapoints, the privatized parameters are closer to those given by the conventional EM algorithm. However, now the mean parameter for the married category provides aggregated information from several people, which makes it hard to infer any individual information.    }
\label{fig:motivatingfig}
\end{figure*}


\section{DPEM for mixture of Gaussians}\label{DPEMMoG}


\subsection{EM for Mixture of Gaussians}
We consider the mixture of Gaussians (\emph{MoG}) model as a first example to derive the DP-EM algorithm. For $K$ Gaussians and $N$ data points $X:=\{ \vx_i \}_{i=1}^N$,  the log-likelihood under MoG is given by 
$\log p(X|\vpi, \vmu, \Sigma)= \sum_{i=1}^N \log \sum_{k=1}^{K} \pi_{k} \Nrm(\vx_i|\vmu_k, \Sigma_k),$
where $\sum_{k=1}^{K} \pi_{k} =1$. We denote the parameters by $\vtheta := \{ \vpi, \vmu, \Sigma\} = \{ \pi_k, \vmu_k, \Sigma_k \}_{k=1}^K$. 

Introducing a binary vector of length $K$ for each data point, $\vz_i \in \mathbb{R}^K$, to represent the membership to which Gaussian each datapoint belongs, e.g., $z_{i,k} \in \{ 0, 1\}$ and $\sum_{k=1}^K z_{i,k} = 1$,  the distribution over each $\vz_i$ is given by
$p(\vz_i) := \prod_{k=1}^K \pi_{k}^{\vz_{i,k}},$
and the distribution over all unobserved variables $Z = \{ \vz_i \}_{i=1}^K$ is given by 
$p(Z) := \prod_{i=1}^N p(\vz_i).$ The joint distribution over observed and unobserved variables, which is in the exponential family, is given by 
$\log p(X, Z|\vpi, \vmu, \Sigma) = \sum_{i=1}^N \sum_{k=1}^{K} z_{i,k} [ \log \pi_{k} + \log \Nrm(\vx_i|\vmu_k, \Sigma_k)].$
In the E-step, we compute the responsibilities as $\langle \delta_{z_{i,k}=k} \rangle_{q(Z)}$ given the parameters from the previous iteration $\vtheta^{prev}$
\begin{align}\label{eq:responsibilities}
\small{\gamma_{i,k}} &= \small{p(z_{i,k}=1|\vx_i, \vtheta^{prev})},\nonumber\\
& = \small{\pi_k \Nrm(\vx_i | \vmu_k, \Sigma_k)/{\sum_{k=1}^K\pi_k \Nrm(\vx_i | \vmu_k, \Sigma_k)}}\normalsize,
\end{align} 
and in the M-step, we update the parameters $\vtheta$ by
\begin{align}\label{eq:paramupdateMoG}
\small{\pi_k^{MLE}}&= \small{\frac{N_k}{N}, \quad \vmu_{k}^{MLE} = \frac{1}{N_k} \sum_{i=1}^N \gamma_{i,k} \vx_i}, \\  \small{\Sigma_{k}^{MLE}}&= \small{\frac{1}{N_k} \sum_{i=1}^N \gamma_{i,k} (\vx_i -\vmu_{k}^{MLE} ) (\vx_i -\vmu_{k}^{MLE} )\trp}\normalsize, \nonumber
\end{align} 
where $N_k = \sum_{i=1}^N  \gamma_{i,k}$. 

For the maximum a posteriori estimate, we impose the Dirichlet prior on $\vpi \sim \mbox{Dir}(\valpha)$ and Normal-inverse-Wishart prior on $p(\vmu_k, \Sigma_k) = \mbox{NIW}(\mathbf{0}, \kappa_0, \nu_0, S_0)$, where the MAP estimates are
\begin{align}
\small{\pi_k^{MAP}} &= \small{\frac{N \pi_k^{MLE} + \alpha_k - 1}{ N + \sum_k \alpha_k -K }} , \quad \small{\vmu_k^{MAP} = \frac{N_k \vmu_k^{MLE} }{N_k + \kappa_0}}, \nonumber \\
\small{\Sigma_k^{MAP}} &= \small{\frac{S_0 + N_k \Sigma_k^{MLE} + \frac{\kappa_0 N_k}{\kappa_0+ N_k} \vmu_k^{MLE}\vmu_k^{MLE}\trp}{\nu_0 + N_k + d + 2}} \mbox{ .} \nonumber
\end{align}
\normalsize
In this paper we set hyperparameters to conventional values, e.g. $\valpha = [2, 2, \cdots, 2], \kappa_0 = 1, \nu_0 = d+2, S_0 =  \mbox{diag}(0.1, \cdots 0.1)$, rather than optimizing them, cf. \cite{Bishop06}. 

Before moving to the next section, we would like to motivate why it is important to construct a privacy preserving algorithm for MoG. In \figref{motivatingfig}, we show that if one runs the EM algorithm for the given dataset, an individual's information can be easily revealed by just looking at the EM parameters, while the noised-up parameters obtained by the method, which will be described next, protect private information effectively.\footnote{We first pre-processed the data by scaling down the magnitude with the maximum L2 norm of the data points, and then added noise to each parameter following the derivations in \secref{DPEMMoG}. For visualisation, we map the results back to the original latitude/longtitude space.}


\subsection{DPEM for MoG}

Under MoG, we plug in the responsibilities given in \eqref{responsibilities} to the parameter update expressions given in \eqref{paramupdateMoG}. We then perturb each of these by taking into account one datapoint's worst-case difference between two neighboring datasets. 
We use $\epsilon_i$ to denote a privacy budget allocated per iteration. 



\paragraph{$\epsilon_i$-DP or $(\epsilon_i, \delta_i)$-DP mixing coefficients.}
For two neighbouring datasets with a single data point difference, the maximum difference in $\vpi$ occurs when the data point $\vx_j$ is assigned to the $k$-th Gaussian with  $\gamma_{j,k}=1$ and the altered data point ${\vx}'_j$ is assigned to another, e.g., the $k'$-th Gaussian, with ${\gamma}'_{j,k'}=1$. Hence, we get the following sensitivity:
\begin{eqnarray}
\Delta \vpi^{MLE}
%
=  \max_{\vx_j, {\vx}'_j} \sum_{k=1}^K \tfrac{1}{N}|  \gamma_{j,k} - {\gamma}'_{j,k}|\leq{2}/{N},
\end{eqnarray}  
since $ 0\leq \gamma_{j,k} \leq 1$ and $\sum_{k=1}^K \gamma_{j,k} = 1$.  We add noise to compensate the maximum difference\footnote{To ensure $ \tilde{\vpi}_k^{MLE} \in [0, 1] $, we set $\tilde{\vpi}_k^{MLE} =0$, if $\tilde{\vpi}_k^{MLE} <0$, and $\tilde{\vpi}_k^{MLE} =1$, if  $\tilde{\vpi}_k^{MLE} >1$. 
Then, we re-normalize $\tilde{\pi}^{MLE}$ after the projection to ensure $\sum_{k=1}^K \tilde{\pi}_k^{MLE} = 1$. }
\vspace{-0.05cm}
\begin{equation}\label{eq:perturbmixing}
\tilde{\vpi}^{MLE} = {\vpi}^{MLE} + (Y_1,\cdots, Y_K), 
\vspace{-0.2cm}
\end{equation} where $Y_{i} \sim^{i.i.d.}$ $\mbox{Lap}(\frac{\Delta \vpi^{MLE}}{\epsilon'})$ or $\Nrm(0, \sigma^2)$ with $\sigma^2 \geq 2\log(1.25/\delta_i)(\Delta \vpi^{MLE})^2/\epsilon_i^2$.
For $\pi_k^{MAP}$, we do not need any additional sensitivity analysis, since the MAP estimate is a deterministic mapping of the MLE. 


\paragraph{$\epsilon_i$-DP  or $(\epsilon_i, \delta_i)$-DP mean parameters.}
Using the noised-up $\tilde{N}_k$ obtained from the noised-up mixing coefficients, i.e., $\tilde{N}_k = N \tilde{\pi}_k$, the maximum difference in mean parameters due to one datapoint's difference is 
\vspace{-0.1cm}
\begin{align}\label{eq:mean_sensitivity}
\small{\Delta_1 \vmu_k^{MLE}}
&= \small{ \max_{\vx_j, {\vx}'_j} \;   \tfrac{1}{ \tilde{N}_k} \left| (A_k+ \gamma_{j,k} \vx_j )  -(A_k+ {\gamma}'_{j,k} {\vx}'_j) \right|_1}, \nonumber \\
& \leq \small{  2{\sqrt{d}}/{\tilde{N}_k}},
\vspace{-0.2cm}
\end{align} \normalsize
where $A_k:=\sum_{i=1, i\neq j}^N \gamma_{i,k} \vx_i$ and the L1 term is bounded by \eqref{moment_sen}. 
The $\sqrt{d}$ term is from the fact that each input vector is L2-norm bounded by 1.\footnote{
$ \sum_{l=1}^d | \vx_{i, l}| \leq \left(\sum_{l=1}^d |\vx_{i,l}|^2\right)^{\frac{1}{2}} \left(\sum_{l=1}^d 1\right)^{\frac{1}{2}} \leq \sqrt{d}$. }
We add noise to the MLE via\footnote{The MAP estimate only differs from the MLE in the denominator. Hence, we simply replace $\tilde{N}_k$ with $\tilde{N}_k+\kappa_0$ in \eqref{mean_sensitivity} in the MAP estimation case.}
\vspace{-0.02cm}
\begin{eqnarray}\label{eq:perturbmean}
\tilde{\vmu}_k^{MLE} = {\vmu}_k^{MLE} + (Y_1,\cdots, Y_{d}),
\end{eqnarray} 
 where $Y_{i} \sim^{i.i.d.}$ $\mbox{Lap}({\Delta_1 \vmu_k^{MLE}}/{\epsilon'})$ or $\Nrm(0, \sigma^2)$ with $\sigma^2 \geq 2\log(1.25/\delta_i)(\Delta_2 \vmu_k^{MLE})^2/\epsilon_i^2$, where $\Delta_2 \vmu_k^{MLE} = 2/\tilde{N}_k$.
 
\paragraph{$(\epsilon_i,\delta_i)$-DP covariance parameters.}
For covariance perturbation, we follow the Analyze Gauss (AG) algorithm \cite{DworkTT014}, which provides $(\epsilon_i, \delta_i)$-DP. 
We first draw Gaussian random variables 
\begin{equation}
\vz \sim \Nrm\left(0, \beta I_{d(d+1)/2}\right), 
\end{equation}
where
$\beta = 2 \log (1.25/\delta_i) (\Delta \Sigma_k^{MLE})^2 / (\epsilon_i)^2 $ and the sensitivity of the covariance matrix
\footnote{
The MAP estimate only differs from the MLE in the denominator. We replace $\tilde{N}_k$ with $\tilde{N}_k + \nu_0 + d + 2$ in \eqref{Cov_sens} in the MAP estimation case.}
 in Frobenius norm is given by 
\begin{align}\label{eq:Cov_sens}
\Delta \Sigma_k^{MLE} &=\max_{\vx_j, \vx'_j}  \tfrac{1}{\tilde{N}_k}|\mbox{vec}\{(B_k + \gamma_{j,k}\vx_j \vx_j\trp-\tilde{M}_k ) \nonumber \\
 & \qquad \qquad \qquad - (B_k + {\gamma}'_{j,k} {\vx}'_j {\vx}'_j\trp -\tilde{M}_k )\}|_2, \nonumber\\
&\leq \tfrac{2}{\tilde{N}_k} \sqrt{ \sum_{l=1}^d \sum_{l'=1}^d (\vx_{j, l} \vx_{j, l'})^2 } \leq \tfrac{2}{\tilde{N}_k}
\end{align} where $B_k:=\sum_{i=1, i\neq j}^N \gamma_{i,k} \vx_i\vx_i\trp$, and $\tilde{M}_k = \tilde{N}_k \tilde{\vmu}_k^{MLE} \tilde{\vmu}_k^{MLE}\trp$.
Using $\vz$, we construct a upper triangular matrix (including diagonal), then copy the upper part to the lower part so that the resulting matrix $Z$ becomes
symmetric. Then, we add this noisy matrix to the covariance matrix 
\begin{equation}\label{eq:perturbcovfull}
\tilde{\Sigma}_k^{MLE} := \Sigma_k^{MLE} + Z.
\end{equation} The perturbed covariance might not be positive definite. In such case, we project the negative eigenvalues to some value near zero to maintain positive definiteness of the covariance matrix.

\paragraph{Combinations of the perturbations.}
Among all the possible combinations of these parameter perturbation mechanisms, we focus on two scenarios. Scenario 1 (which we call {\it{LLG}}) uses the $\epsilon_i$-DP Laplace mechanism for perturbing mixing coefficients (once) and mean parameters (K times) and the $(\epsilon_i, \delta_i)$-DP Gaussian mechanism for perturbing the covariance parameters (K times). Since there are $K$ Gaussians,  for $J$ iterations, there will be $J(K+1)$ compositions of $\epsilon_i$-DP mechanism and $JK$ compositions of $(\epsilon_i, \delta_i)$-DP mechanisms in total in this scenario. Scenario 2 (which we call {\it{GGG}}) uses the $(\epsilon_i, \delta_i)$-DP Gaussian mechanism for perturbing all the parameters. For $J$ iterations, there will be $J(2K+1)$ compositions of $(\epsilon_i, \delta_i)$-DP mechanism in total in this scenario. 

%
%

\section{Compositions for DP-EM for MoGs}\label{CDPEM}

Before describing our method, we first describe the two baseline methods. First, in {\it{Linear}} (Lin) composition (Theorem 3.16 \cite{Dwork14}), privacy degrades linearly with the number of iterations. This result is from the Max Divergence of the privacy loss random variable being bounded by a total budget. Hence, the linear composition yields ($J(2K+1)\epsilon_i$, $JK\delta_i$)-DP under scenario $LLG$ and ($J(2K+1)\epsilon_i$, $J(2K+1)\delta_i$)-DP under scenario $GGG$. Second, {\it{Advanced}} (Adv) composition (Theorem 3.20 \cite{Dwork14}), resulting from the Max Divergence of the privacy loss random variable being bounded by a total budget including a slack variable $\delta$, yields
$(J(2K+1) \epsilon_{i} (e^{\epsilon_{i}}-1) + \sqrt{2J(2K+1)\log(1/\delta')}\epsilon_{i}, \; \delta' + J K \delta_i)$-DP
under scenario $LLG$
and 
$(J(2K+1) \epsilon_{i} (e^{\epsilon_{i}}-1) + \sqrt{2J(2K+1)\log(1/\delta')}\epsilon_{i}, \; \delta' + J(2K+1) \delta_i)$-DP
under scenario $GGG$.

Our method calculates the per-iteration budget using the two composition methods below. 
\paragraph{zCDP composition (zCDP).}
{\it{z-CDP}} composition
yields $(\rho + 2 \sqrt{\rho \log (1/\delta)}, \delta)$-DP, where 
\begin{align}
\rho = J(K+1) {\epsilon_i^2}/{2} + JK\Delta \Sigma_k^2/(2\sigma_3^2) \nonumber
\end{align} under scenario $LLG$
and 
\begin{align}
\rho = J\Delta \vpi ^2/(2\sigma_1^2) + JK\Delta \vmu_k^2/(2\sigma_2^2) + JK\Delta \Sigma_k^2/(2\sigma_3^2)\nonumber
\end{align} under scenario $GGG$, 
%
%
for sensitivity $\Delta \vpi, \Delta \vmu_k, \Delta \Sigma_k$ and $\sigma_1^2 \geq 2 \log(1.25/\delta_i) {\Delta \vpi^2}/{\epsilon_i^2}$, $\sigma_2^2 \geq 2 \log(1.25/\delta_i) {\Delta \vmu_k^2}/{\epsilon_i^2}$, and $\sigma_3^2 \geq 2 \log(1.25/\delta_i) {\Delta \Sigma_k^2}/{\epsilon_i^2}$,  where $0<\epsilon_i<1$.

These results are obtained by using the following results in \cite{BunS16}: Proposition 1.4. If $\mathcal{M}$ satisfies $\epsilon_i$-DP, then  $\mathcal{M}$ satisfies $\frac{1}{2}\epsilon_i^2$-zCDP; Proposition 1.6. Gaussian mechanism satisfies ${\Delta^2}/({2\sigma^2})$-zCDP, where $\Delta$ is a sensitivity; 
Lemma 1.7. If two mechanisms satisfy $\rho_1$-zCDP and $\rho_2$-zCDP, respectively, then their composition satisfies $(\rho_1+\rho_2)$-zCDP; and Proposition 1.3. If $\mathcal{M}$ provides $\rho$-zCDP, then $\mathcal{M}$ is $(\rho + 2 \sqrt{\rho \log (1/\delta)}, \delta)$-DP for any $\rho>0$. 

\paragraph{Moments Accountant composition (MA).}
For using MA, as a first step, we identify the form of privacy loss random variable and its $\lambda$-th moment in each mechanism we use. 
For $\epsilon_i$-DP Laplace mechanism $\mathcal{M}^{L}_i$ outputting $f(\Dat)$ and  
$x \sim \mbox{Lap}(0, \frac{\Delta f}{\epsilon_i})$,  $L^{(o)}$ at $o = f(\Dat) + x$ has the following form:
\begin{align}
L^{(o)} = 
\begin{cases}
\epsilon_i, \; \mbox{if $x<$ 0}, \; \mbox{w.p. }\frac{1}{2} \\
- \epsilon_i, \; \mbox{if $x >\Delta f$}, \; \mbox{w.p. }\frac{1}{2} \mbox{e}^{-\epsilon_i}\\
-\frac{\epsilon_i}{\Delta f} (2x - \Delta f), \; \mbox{if $ 0 \leq x \leq \Delta f$}, \; \mbox{w.p.}\frac{1}{2} (1-\mbox{e}^{-\epsilon_i}). \nonumber 
\end{cases}
\end{align}
Following the definition in \eqref{Moment}, the $\lambda$-th moment is given by 
\vspace{-0.2cm}
\begin{align}
\label{eq:Moment_Laplace}
\alpha_{\mathcal{L}}
= \log \left[ \frac{\lambda + 1}{2 \lambda + 1} e^{\lambda \epsilon_i} + \frac{\lambda}{2\lambda + 1}e^{(- \epsilon_i (\lambda + 1))} \right].
\vspace{-0.2cm}
\end{align}
For $(\epsilon_i, \delta_i)$-DP Gaussian mechanism $\mathcal{M}^{G}_i$
with noise magnitude $\sigma$ and $x \sim \Nrm(0, \sigma^2)$,  $L^{(o)}$ at $o = f(\Dat) + x$ is
$L^{(o)}= \left(\frac{\Delta f}{\sigma} \right) \left(\frac{x}{\sigma} \right) + \frac{1}{2} \left(\frac{\Delta f}{\sigma} \right)^2.$
%
The $\lambda$-th moment is then
\vspace{-0.2cm}
\begin{align}
\label{eq:Moment_Gaussian}
\alpha_{\mathcal{G}} 
= (\lambda^2 + \lambda) \frac{(\Delta f)^2}{2 \sigma^2}.
\vspace{-0.2cm} 
\end{align} 
Note that multi-dimensional Laplace/Gaussian mechanisms also have the same form of the $\lambda$-th moment as the scalar version. See the Supplementary material for the derivation.

For achieving ($\epsilon, \delta$)-DP, the tail bound is given by  
$ \delta 
 =   \min_{\lambda} \exp \left[ J(K+1) \alpha_{\mathcal{L}} + JK\alpha_{\mathcal{G}}  - \lambda \epsilon \right]$ under scenario $LLG$; and  
 $ \delta 
 =   \min_{\lambda} \exp \left[ J(2K+1) \alpha_{\mathcal{G}} - \lambda \epsilon \right]$ under scenario $GGG$. Under each case, 
 we calculate $\epsilon_i$ satisfying the tail bound with the fixed budget $(\epsilon,\delta)$. 
Algorithm \ref{algo:DPEM_MoG} summarizes our method. 
\begin{algorithm}[h]
\caption{DP-EM under MoG using MA}\label{algo:DPEM_MoG}
\begin{algorithmic}
\vspace{0.1cm}
\REQUIRE Dataset $\Dat$, per-iteration budget ($\epsilon_i$, $\delta_i$) calculated by MA or zCDP composition
\vspace{0.1cm}
\ENSURE ($\epsilon, \delta$)-DP parameters $\tilde{\vtheta}$\\
\vspace{0.1cm}
\textbf{Iterate until convergence (J iterations):}
\vspace{0.1cm}
\STATE Compute parameters by plugging in the\\ responsibilities given in \eqref{responsibilities}.
\STATE Noise up $\vpi$ by \eqref{perturbmixing}, $\vmu$ by \eqref{perturbmean}, and \\$ \Sigma$ by \eqref{perturbcovfull}.
\end{algorithmic}
\end{algorithm}

\section{DPEM for Factor Analysis}\label{FA}



Under FA, the conditional distributions over observed variables $\vx_i$ are assumed to be Gaussian, $p(\vx_i|\vz_i) = \Nrm(\vx_i|W\vz_i, \Psi)$, and the prior over latent variables $\vz_i$ is also assumed to be Gaussian: $p(\vz_i) = \Nrm(\vz_i|0, I)$.

In this case, the complete-data likelihood is proportional to
$p(X, Z) \propto \exp(\phi(\vtheta)\trp T(X, Z))$,
where $\phi(\vtheta)$ is a vectorized version of the concatenated matrix $[ W\trp \Psi^{-1}, \Psi^{-1}, -\frac{1}{2} G^{-1}]$, $G^{-1} = I+W\trp \Psi^{-1}W$, and
where the sufficient statistics are also a vectorized version of a concatenated matrix $T(X, Z) = [ \sum_{i=1}^N\vx_i \vz_i\trp, \sum_{i=1}^N \vx_i \vx_i\trp, \sum_{i=1}^N \vz_i \vz_i\trp ].$

Due to conjugacy the posterior over $\vz_i$ is also Gaussian, where the first and second moments are given by $\bar{\vz}_i = G W\trp \Psi^{-1} \vx_i$ and $\langle\vz_i \vz_i\trp\rangle = G + \bar{\vz}_i \bar{\vz}_i\trp$. The expected sufficient statistics become a function of the data second moment matrix, denoted by $\Lambda := \frac{1}{N} \sum_{i=1}^N \vx_i \vx_i\trp$,
\vspace{-0.2cm}
\begin{align} 
& \langle T(X, Z) \rangle_{q(Z)}\nonumber \\
 &= N \left[  \Lambda \Psi^{-1} W G\trp , \; \Lambda, \; G + GW\trp\Psi^{-1}\Lambda\Psi^{-1}WG\trp \right]. \nonumber
 \vspace{-0.2cm}
\end{align} 
For privacy-preserving EM, we perturb $\Lambda$ by Analyze Gauss \cite{DworkTT014}, resulting in a perturbed matrix $\tilde{\Lambda}$, which we use when updating the parameters by 
\vspace{-0.2cm}
\begin{align}
W^{new} &= \left[ \tilde{\Lambda} \Psi^{-1} W G\trp \right]  \left[ G + G W\trp \Psi^{-1} \tilde{\Lambda} \Psi^{-1} W G\trp \right]^{-1}, \nonumber \\
\Psi^{new} &= \mbox{diag} \left[ \tilde{\Lambda} - W G W\trp \Psi^{-1} \tilde{\Lambda} \right].\nonumber
\vspace{-0.2cm}
\end{align} until convergence, at no extra privacy cost. 
Therefore, unlike MoGs, FA only requires perturbing the data second moment matrix {\it{once}} for privacy preservation. The EM iterations are then \emph{post-processing} steps which are free from cumulative differential privacy loss.  

\section{Experiments}\label{Illustration}

We used four real-world datasets to test our algorithm. In all datasets, we preprocessed the data such that the input vectors had maximum norm 1.

\begin{figure}[t]
\centering
\centerline{\includegraphics[width=0.4\textwidth]{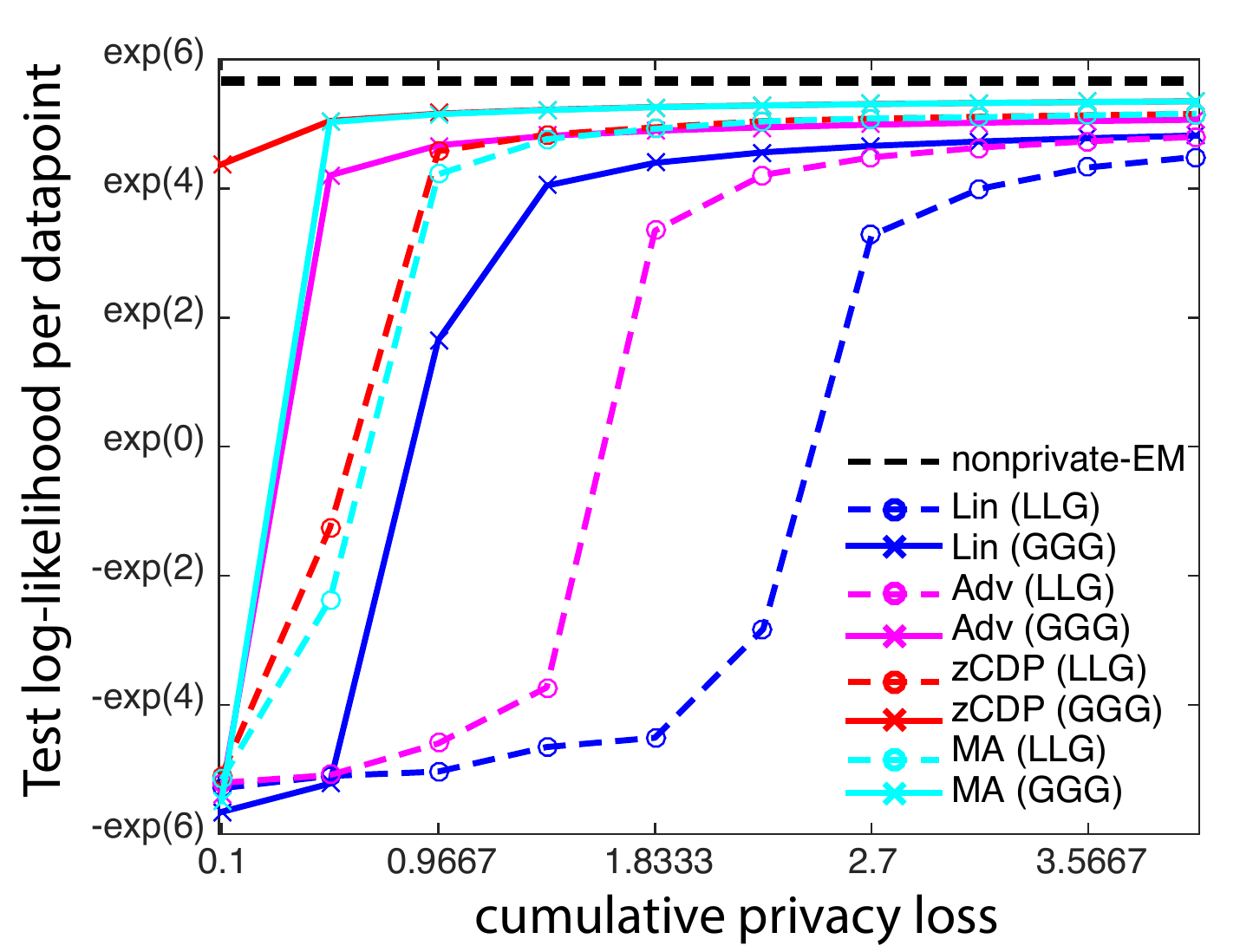}}
  \caption{\textbf{Stroke dataset}. Test log-likelihood per data point as a function of cumulative privacy loss after $20$ EM iterations. We fit the data with MoG using the conventional EM first (in black dotted line). We then ran the private EM algorithm with a different per-iteration privacy budget resulting from different composition methods, in order to achieve ($\epsilon, \delta$)-DP EM parameters, where $\epsilon$  varies from $0.1$ to $4$ and $\delta$ is fixed  to $10^{-4}$. We fixed $\delta_i=10^{-6}$  when using Gaussian mechanisms. 
}
\label{fig:Stroke}
\end{figure}

\textbf{Stroke dataset}
was used in \cite{LethamRuMcMa14} for predicting the occurrence of a stroke within a year after an atrial fibrillation diagnosis. 
We used $100$ principal components ($d=100$) of $4,096$ raw features (conditions and medicines) recorded from $50,345$ patients, by assuming that the private database was given in this form. 
We divided the extracted dataset into 10 different pairs of training ($90 \%$) and test sets ($10 \%$), and reported the average test log-likelihood per datapoint across the 10 independent trials in \figref{Stroke}, setting $k=10$.

Overall, the \emph{GGG} scenario yielded higher test log-likelihoods than the \emph{LLG} scenario, so we focused on this method in our experiments. We found that using the zCDP and MA compositions resulted in more accurate estimates, while also requiring less privacy budget, compared to other compositions. zCDP performed better than MA with a small privacy budget $\epsilon$, but they both performed similarly well with a larger budget. The difference with small $\epsilon$ may be due to only searching over integer values of $\lambda$ for MA, which we do for computational reasons, following \cite{2016arXiv160700133A}.

\begin{figure}[t]
\centering
\includegraphics[width=0.4\textwidth]{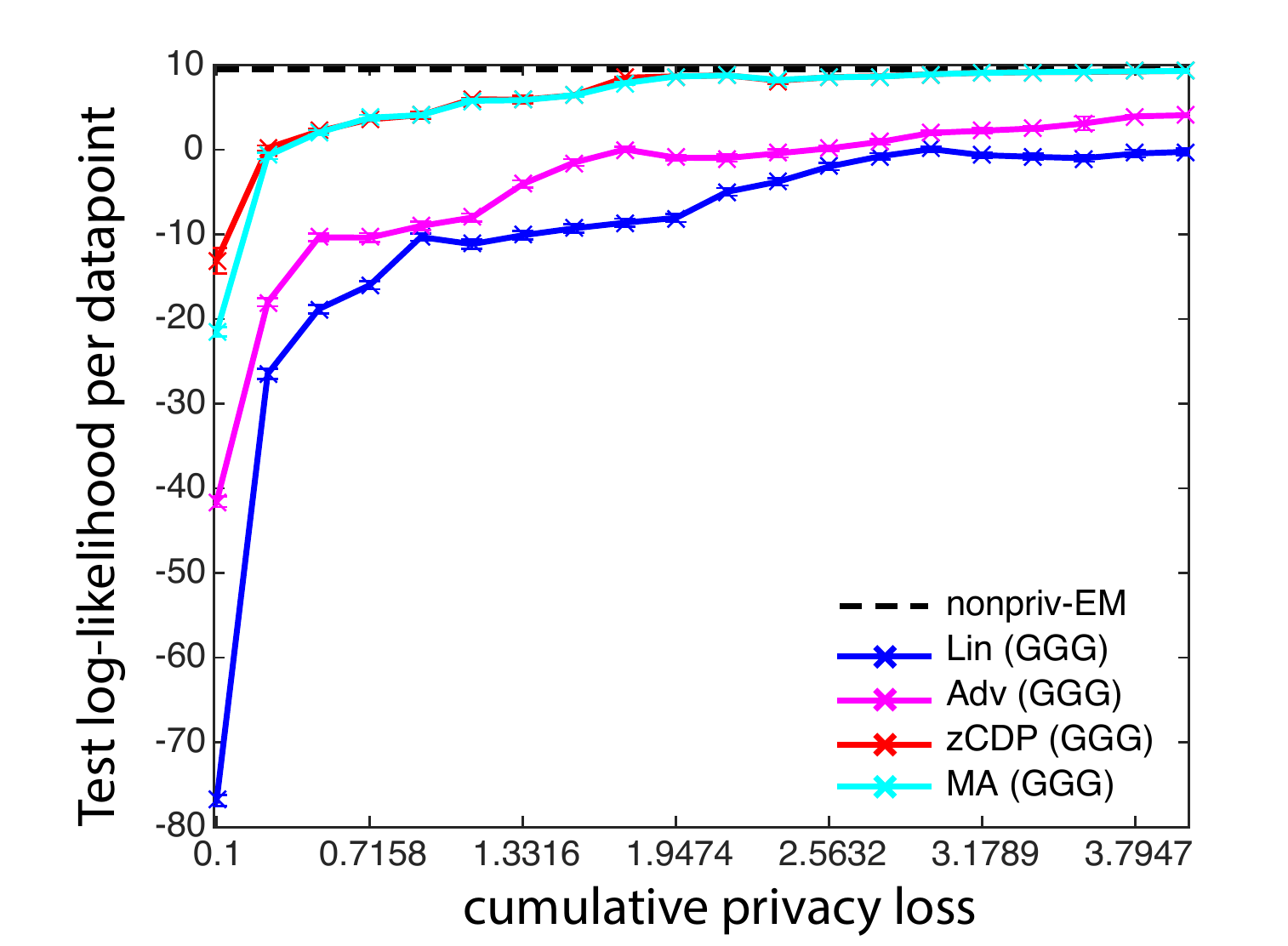}
  \caption{\textbf{Life Science dataset} Test log-likelihood per data point as a function of cumulative privacy loss after $10$ EM iterations. We fit the data with MoG using the conventional EM first (in black dotted line). We then ran the DP-EM algorithm (GGG combination) with a different per-iteration privacy budget resulting from different composition methods, in order to achieve ($\epsilon, \delta$)-DP EM parameters, where $\epsilon$  varies from $0.1$ to $4$ and $\delta$ is fixed  to $10^{-4}$. We fixed $\delta_i=10^{-8}$. 
}
\label{fig:Lifesci}
\end{figure}

\textbf{Life Science dataset}
is from the UCI repository \cite{Lichman:2013}. The dataset contains 26,733 records, consisting of 10 principal components from chemistry and biology experiments ($d=10$). Following other approaches (e.g., \cite{MohanTSSC12}), we set $k=3$. We divided the dataset into 10 different pairs of training ($90 \%$) and test sets ($10 \%$), and reported the average test log-likelihood per data point across the 10 independent trials in \figref{Lifesci}. In this experiment, we focused on scenario $GGG$. Using the zCDP and MA compositions once again resulted in more accurate estimates while requiring less privacy budget than linear and advanced compositions.

\begin{figure*}[t]
\centering
\centerline{\includegraphics[width=0.9\textwidth]{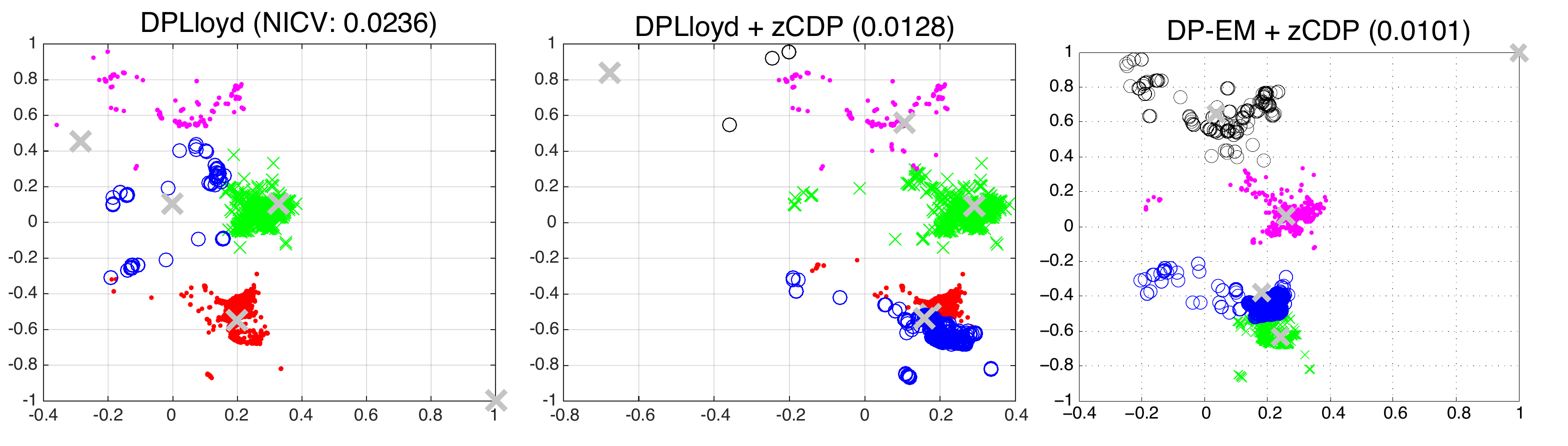}}
  \caption{\textbf{k-Means Clustering.} Visualisation of clustering results with total privacy budget $\epsilon = 0.01$ and tolerance  $\delta=10^{-4}$. The center locations are depicted in gray cross. The numbers in parenthesis are \emph{normalised intra-cluster variance} (NICV) values obtained by each method. 
 \textbf{Left} The DPLloyd algorithm with linear composition performed poorly due to the relatively high level of additive noise.   
 \textbf{Middle} DPLloyd with zCDP composition performed better than the original version.
   \textbf{Right} Our algorithm achieves smaller NICV than two variants of DPLloyd given the same privacy budget. 
  }
\label{fig:Gowalla_kMeans}
\end{figure*}

\begin{figure*}[t]
\centering
\centerline{\includegraphics[width=1\textwidth]{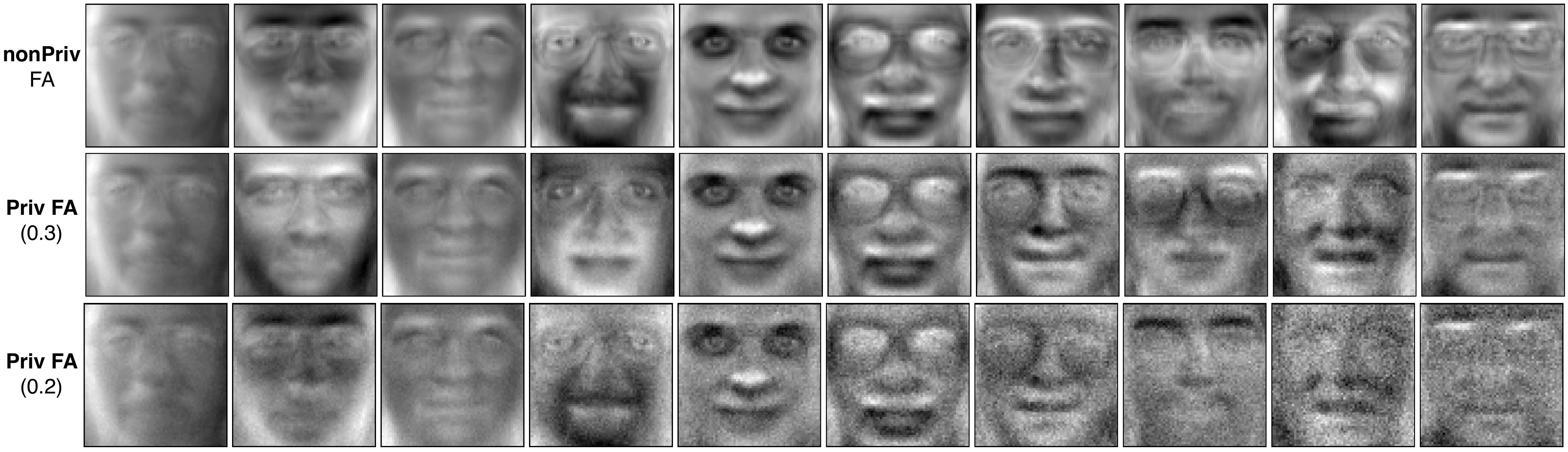}}
  \caption{\textbf{Private Factor Analysis.}  Visualisation of each column of estimated $W$ (reshaped as $64$ by $64$ images).
  \textbf{Top}: Non-private EM.
  \textbf{Middle}: DP-EM with $\epsilon = 0.3$ and $\delta=10^{-4}$. 
 \textbf{Bottom}:  DP-EM with $\epsilon = 0.2$ and $\delta=10^{-4}$. 
  }
\label{fig:FA}
\end{figure*}

\textbf{Gowalla dataset} 
 contains the social network's users' check-in locations in terms of longitude and latitude ($d=2$). 
The total number of data points  is 1,256,384, which we divided into $10$ cross-validation sets. 
We then performed $k$-means clustering and compared our method to a differentially private $k$-means clustering algorithm, DPLloyd  \cite{Blum:2005:PPS:1065167.1065184}. 
The standard Lloyd algorithm for $k$-means clustering first partitions the data into $k$ clusters, with each point assigned to be in the same cluster as the nearest centroid, and then updates each centroid to be the center of the data points in the cluster. 
 As summarized in \cite{Su2016}, the DPLloyd adds noise to the updated centroids. Specifically, the Laplace noise is added to the number of data points assigned to each cluster as well as to the sum of each coordinate of the data points assigned to each cluster. Hence, the sensitivity becomes $d+1$.  
In the original DPLloyd algorithm,  due to the conventional composition theorem for DP, their noise distribution follows Lap$\left({(d+1)J}/{\epsilon}\right)$ for $J$ iterations. We also tested the DPLloyd algorithm with zCDP compositions, which resulted in better performance in terms of normalized intra-cluster variance (NICV) across the $10$ test sets. 
Our algorithm for $k$-means clustering also perturbs the centroids by adding the Laplace noise with zCDP composition, where the sensitivity of the mean locations is given in \eqref{mean_sensitivity}.
We set $\epsilon=0.01$ and $\delta=10^{-4}$ for both algorithms.  
As shown in \figref{Gowalla_kMeans}, our method achieves smaller NICV than DPLloyd, even with a very small value of $\epsilon$.

\textbf{Olivetti Faces dataset} 
is used to illustrate our private factor analysis method\footnote{We obtained the dataset from \url{http://scikit-learn.org/}, but the dataset is originally from $AT\&T$ Laboratories Cambridge.}. The dataset consists of  ten different images for each of 40 distinct subjects ($N=400$), where each image is 64 by 64, resulting in $4096$ features ($d=4096)$. 
Each pixel is a floating point value on the interval $[0, 1]$.
Each image was treated as a datapoint, rather than each subject, though this could readily be done via \emph{group privacy} \cite{Dwork14}.
We set the latent dimension to $10$. We tested non-private EM, DP-EM with $\epsilon=0.2$ and $\epsilon=0.3$ (fixing $\delta=10^{-4}$), and showed each column of the estimated loading matrix $W$ in \figref{FA}. With $\epsilon = 0.2$ (bottom) the components were noisy, but with $\epsilon = 0.3$ (middle) the FA components' faces were nearly as recognizable as for the non-private FA algorithm (top), thereby accurately recovering a set of typical faces in the dataset.

\section{Conclusion}

We have developed a practical algorithm that outputs accurate and privatized EM parameters based on moment perturbation under the MA and zCDP composition analyses, which effectively decrease the amount of additive noise for the same expected privacy guarantee compared to the standard analysis. 
We illustrated the effectiveness of our algorithm on four datasets. Based on our results, we recommend the use of zCDP composition analysis for EM, since it performed better than MA in some regimes and is easier to compute. Furthermore, we found that the GGG combination performed better than LLG under these composition methods in the context of EM, which perhaps makes sense since the zCDP and MA compositions are tailored to the Gaussian mechanism. 


The private EM algorithms for the mixture of Gaussians and factor analysis models we discussed in this paper are clearly only two examples of a much broader class of models to which our private EM framework applies. Our positive empirical results with EM strongly suggest that these ideas are likely to be beneficial for privatizing many other iterative machine learning algorithms. 
In future work, we plan to apply this general framework to other inference methods. This fits our broader vision that \emph{practical} privacy preserving machine learning algorithms will have an increasingly relevant role to play in our field.

\newpage
\small
\bibliographystyle{unsrt}
\bibliography{DPEPrefs}

\end{document}